\documentclass[10pt,twocolumn,letterpaper]{article}

\usepackage[pagenumbers]{iccv} %

\usepackage{booktabs}
\usepackage{multirow}
\usepackage{rotating}
\usepackage{adjustbox}
\usepackage{bm}
\usepackage{soul}
\def\eg{\emph{e.g}\onedot}

\def\ie{\emph{i.e}\onedot}

\usepackage[accsupp]{axessibility}  %
\newcommand{\custompar}[1]{
  \par
  \vspace{4pt}
  \noindent\textbf{#1}
}

\definecolor{iccvblue}{rgb}{0.21,0.49,0.74}
\usepackage[pagebackref,breaklinks,colorlinks,allcolors=iccvblue]{hyperref}

\def\paperID{14356} %
\def\confName{ICCV}
\def\confYear{2025}

\title{AstroLoc: Robust Space to Ground Image Localizer}

\author{Gabriele Berton$^{1,2^*}$
\quad
Alex Stoken
\quad
Carlo Masone$^{1}$\\
$^{1}$Politecnico di Torino 
$^{2}$Amazon
\\
}

\begin{document}
\makeatletter
\def\@maketitle{
  \newpage
  \null
  \vskip 2em
  \begin{center}
    {\LARGE \@title \par}
    \vskip 1.5em
    {\large
      \lineskip .5em
      \begin{tabular}[t]{c}
        \@author
      \end{tabular}\par}
    \vskip 1em
    {\url{https://astro-loc.github.io/}}
  \end{center}
  \par
  \vskip 1.5em
}
\makeatother
\maketitle

\begin{abstract}

Thousands of photos of Earth are taken every day by astronauts from the International Space Station. Localizing these photos, which has been performed manually for decades, has recently been approached through image retrieval solutions: given an astronaut photo, the goal is to find its most similar match among a large database of geo-tagged satellite images, in a task called Astronaut Photography Localization (APL). Yet, existing APL approaches are trained only using satellite images, without taking advantage of the millions of open-source astronaut photos. In this work we present the first APL pipeline capable of leveraging astronaut photos for training. We first produce full localization information for 300,000 manually weakly-labeled astronaut photos through an automated pipeline, and then use these images to train a model, called AstroLoc. AstroLoc learns a robust representation of Earth's surface features through two objective functions: pairing astronaut photos with their matching satellite counterparts in a pairwise loss, and a second loss on clusters of satellite imagery weighted by their relevance to astronaut photography through unsupervised mining. AstroLoc achieves a staggering 35\% average improvement in recall@1 over previous SOTA, reaching a recall@100 consistently over 99\% for existing datasets. Moreover, without fine-tuning, AstroLoc provides excellent results for related tasks like the lost-in-space satellite problem and historical space imagery localization.
\end{abstract}

\section{Introduction}
\label{sec:intro}

\begin{figure}
    \begin{center}
    \includegraphics[width=0.99\columnwidth]{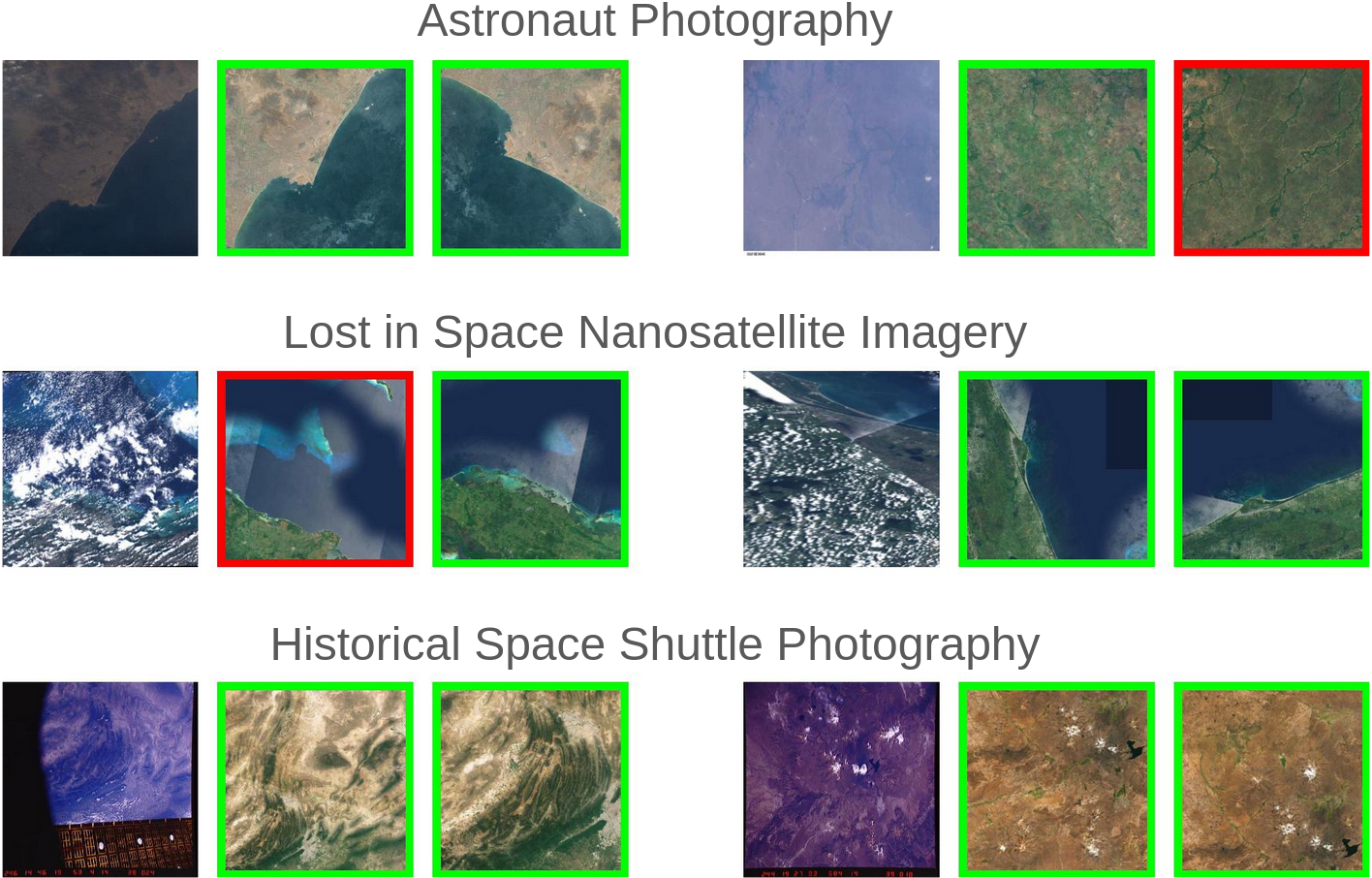}
    \end{center}
    \caption{\textbf{One model, many space to ground applications.} We train a single model, AstroLoc, that succeeds in multiple space-based image retrieval settings: astronaut photography localization, “lost in space” orbit determination, and historical (Space Shuttle) photography localization.
    In this figure, each group of 3 images represent a query and its top-2 predictions from searching over a worldwide database of millions of satellite images. Correct predictions in green, wrong predictions in red.
    }
    \label{fig:teaser}
\end{figure}

\begingroup
\renewcommand\thefootnote{}
\footnotetext{*Work was done outside of Amazon}
\endgroup

Earth observation is the process of collecting information about the Earth's surface, which is vital for monitoring the state of our planet.
Among the multitude of systems used to acquire Earth observations data, most of which rely on automatic collection by nadir-facing (straight-down) satellites,
manual acquisition of photos by astronauts aboard the International Space Station (ISS) is a clear outlier.
This imagery has distinctive properties including (1) a wide range of spatial resolutions (up to 2 meters per pixel), (2) the potential for oblique perspectives to observe topography and height (\eg, measure the height of a volcanic eruption to divert flights), (3) various illumination conditions (whereas satellite imagery is typically sun-synchronous, always capturing the same area at the same time of day), and (4) offers the highest resolution open source Earth observations data.
Furthermore, due to the ISS's 90 minute repeating orbit, astronauts can be tasked with imaging for near real time disaster response, and they can use human intuition to focus on relevant areas and events, making the photographs very information-dense, whereas satellites commonly collect images systematically, regardless of semantic value.
All these characteristics make this data source a perfect complement to satellite data and a crucial component for climate science \cite{astro_photos_climate_patterns}, atmospheric phenomena \cite{sprites, TLEs}, urban planning \cite{Sanchez_2022_artificial_lightning, Gaston_2022_environ_impacts_artificial_light, Small_2022_spectrometry_urban_lightscape, Schirmer_2019_nightlight_behavior}, and, most importantly, disaster management and response \cite{IDC_Stefanov}\footnote{\url{https://storymaps.arcgis.com/stories/947eb734e811465cb0425947b16b62b3}}.

Although over 5 million photos have been collected from the ISS since the beginning of its operations, unfortunately their full potential has yet to be unlocked.
The complication is that astronaut photos, unlike satellite imagery, are not automatically geolocated \wrt the Earth's surface.
Even though the position of the ISS can be determined from the timestamp of the photo and the ISS orbit path, an astronaut can point their camera (a standard hand-held DSLR) toward any location on Earth within their vast visibility range, which spans roughly 20 million square kilometers (sqkm)\footnote{Considering that the ISS orbits at an altitude of 415 kilometers.}.
Therefore, localizing a photo that covers an area of 100 sqkm (\ie 0.0005\% of the visible/search area) is like finding a needle in a haystack.
For example, when an astronaut is orbiting above Texas and photographs a wildfire, the burning area could be anywhere from Canada to Mexico depending on camera orientation, and the only way to know where is by geolocating the image.

The potential value these photos hold when geolocated has led to a concerted manual localization effort, with experts and citizen scientists localizing over 300,000 of these images -- a process defined by NASA as a ``monumentally important, but monumentally time-consuming job"\footnote{We highly encourage the reader to view this clip \url{https://www.youtube.com/watch?v=drrP_Iss0gA&t=295s}}. Nevertheless, manually geolocating a single astronaut photo can take hours (hundreds of thousands of human-hours have been spent on hand-labeling 300k of them), effectively making this procedure inadequate given the ever-growing number of images that are being collected (up to 10k per day).
This problem has prompted recent investigation into automating the task of \textbf{Astronaut Photography Localization (APL)} with deep learning, combining image retrieval and image matching techniques~\cite{Stoken_2023_CVPR, Berton_2024_EarthMatch, Stoken_2024_CVPR}.
Image retrieval is used as a first step, to search within a database of precisely geo-referenced satellite images for those that are most similar to the astronaut photo (called the \textit{query}). With such a system, a query can be coarsely localized in less than a second, making it a viable solution for such a large scale problem. Afterwards, the retrieval candidates may be refined using more computationally demanding matching techniques.
Although these pipelines have shown satisfactory speeds for APL, their performance show a wide margin for improvement, and we hypothesize that this is due to existing APL models being trained \textit{only} on satellite images~\cite{Berton_2024_EarthLoc}. This is a critical limitation, considering that the end goal is to geo-reference astronaut photos, not satellite images.

\begin{figure}
    \begin{center}
    \includegraphics[width=0.8\columnwidth]{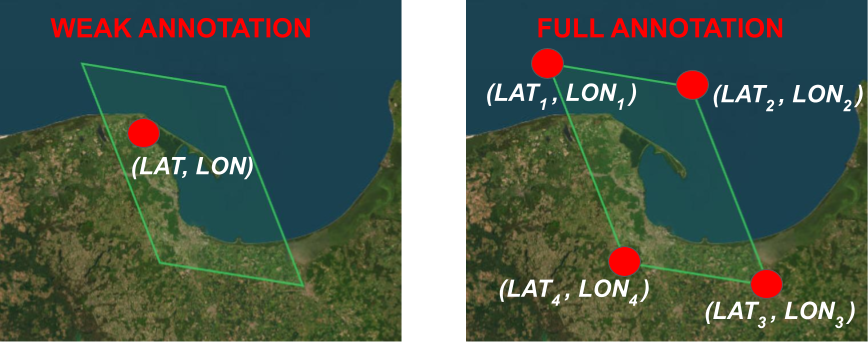}
    \end{center}
    \vspace{-3mm}
    \caption{\textbf{Visual example of weak (manual) annotation and full annotation of an astronaut photo.} Weak annotation is the geographic coordinates of a single point, which does not provide information about the image's size (\ie it could cover a town or an entire continent), whereas full annotation provides coordinates for all 4 corners (called \textit{footprint}), from which the coordinate of any pixel within the image can be easily calculated (pixel-wise label).
    }
    \vspace{-3mm}
    \label{fig:weak_full_annotation}
\end{figure}

We hypothesize that it would be beneficial to exploit the available 300,000 hand-labeled open-source astronaut photos to train the image retrieval model. The difficulty with doing so is that these photos are only weakly annotated, \ie, the manual label only provides the geographic location (latitude,longitude) of a random point close to the center of the image.
Therefore, these weakly labeled photos cannot be directly used in a metric learning framework, where the model must be provided with pairs of queries (astronaut photos) and database (satellite) images from the same location. In order to meet this criteria, we need these photos to be labeled with their full footprint (\ie precise GPS location of its four corners), so that we can exactly determine which images from the database can be used as positive samples (\ie, if they have a substantial overlap \wrt the training queries). The concept of weak and full labels is visualized in \cref{fig:weak_full_annotation}.

In light of these considerations, we first produce a precise annotation of these 300,000 photos, leveraging their weak annotation and a state-of-the-art matching pipeline to label their full footprint.
Then, we demonstrate the importance of using this newly annotated data for training in two ways.
First, we create pairs of matching astronaut-satellite images, and use them in a pairwise contrastive loss that takes advantage of the fact that the two images within each pair come from different domains. The creation of these pairs relies on our new annotations, \ie, we select a pair if the two images have enough intersection over union on Earth's surface.
Even with this loss that takes advantage of the newly labeled 300,000 astronaut photos, it is still important to fully leverage the much larger set of all 5.3M geo-referenced satellite images (not just the ones overlapping astronaut photos). However, we observe that the satellite images are uniformly distributed on the surface of the Earth, whereas the photos from astronauts tend to be unevenly distributed, with higher concentrations in salient areas like volcanoes, glaciers and lakes.
Therefore, we propose a new mining technique, which allows for a weighted sampling of the large satellite image collection according to the \textit{distribution} of astronaut photos, which we call Unsupervised Mining, and, to the best of our knowledge, is the first mining technique of its kind. We then make of use of this data with a second contrastive loss.

We find that our training pipeline, yielding a model which we call \textbf{AstroLoc}, leads to a model so powerful that it saturates existing test sets (recall@100 above 99\%), and brings us to propose new, more challenging test sets that better reflect the task of APL in the real world.
Furthermore, AstroLoc is useful in the real world, and it has already been used to localize hundreds of thousands of images available here\footnote{Photos available at: \url{https://eol.jsc.nasa.gov/ExplorePhotos/}
}. Thanks to AstroLoc, in a few months the backlog of non-localized astronaut photos will be nearly empty for the first time since the launch of the ISS, with the exception of a small fraction that still result in failure cases.
Finally, we note that AstroLoc also shows great potential for other applications like "lost-in-space" satellite orbit determination and historical space image localization.

\section{Related Work}
\label{sec:related_work}
\custompar{Astronaut and satellite imagery localization}
The task of localizing photos taken from Earth's orbit has been studied in a number of domains.
Straub and Christian \cite{Straub_2015_sat_loc} examine the possibility of matching the coastline of imagery from satellites to enhance their autonomous navigation capabilities. Schockley and Bettinger \cite{Shockley_2021_sat_loc} take a similar approach, but instead propose using terrestrial illumination matching.
The problem is more recently addressed by McCleary et al. \cite{McCleary_2024_vinsat}, which selects a set of landmarks on Earth to achieve the same goal. They note that common hardware solutions for localization are expensive ($\sim$10k\$) for nanosatellite developers, making a software solution a game changer for the industry.
For the similar task of localizing astronaut photography, which was historically performed by hand \cite{Stoken_2023_CVPR}, Stoken and Fisher proposed an automated solution leveraging image matching \cite{Stoken_2023_CVPR}, in a framework that was then expanded to localize night-time imagery \cite{Stoken_2024_CVPR}.

While these solutions have advanced the field of Earth from space localization, they are limited, as the matching-based methods can not achieve the scalability and speed required to localize astronaut photographs in real time. Berton et al. \cite{Berton_2024_EarthLoc} proposes approaching the task as an image retrieval problem, where the database is composed of satellite images of known location. This leads to EarthLoc, the first APL-focused retrieval model.
In addition, AnyLoc \cite{Keetha_2023_AnyLoc}, a universal place recognition model, has been shown to produce strong performance and robustness to the query-database domain gap.

\custompar{Unmanned aerial vehicle localization}
A closely related task is that of unmanned aerial vehicle (UAV) localization. Similar to Astronaut Photography Localization (APL), it is approached through cross-domain image retrieval, where the database is often made of satellite imagery and the queries are photos collected by a drone.
The goal is to efficiently find the best prediction for a given query in order to localize the drone in real time \cite{Zheng_2023_UAV_workshop}.
Among notable examples from the literature, 
Bianchi and Barfoot~\cite{bianchi2021uav} introduced a method utilizing autoencoded satellite images, significantly reducing storage and computation costs while maintaining robust localization performance. Dai et al.~\cite{dai2022finding} proposed an end-to-end framework, FPI, that directly identifies UAV locations by matching UAV-view images with satellite-view counterparts, streamlining the localization process. Li et al.~\cite{li2024transformer} developed a transformer-based adaptive semantic aggregation method, whereas He et al. \cite{He_2023_foundloc} make use of foundation models to get an estimate of the location to then refine it with visual-inertial odometry algorithms.

Further improvements in UAV localization are achieved by orientation-guided methods, as shown by Deuser et al.~\cite{Deuser_2023_ogcl_uav_view}, who use orientation-guided contrastive learning to improve UAV-satellite image alignment. Zhu et al.~\cite{Zhu_2023_uav_backbone_winnerUAV_mbeg} proposed a modern backbone architecture optimized for efficient UAV geo-localization, enhancing both speed and accuracy for UAV applications.

\section{Data}
\label{sec:datasets}

Our goal is to localize the extent of \textbf{astronaut photos \textit{(queries)}} given a worldwide map of geo-referenced \textbf{satellite images \textit{(database)}}.
Astronaut photographs represent one of the most diverse and long standing Earth observations data sources. The dataset contains a wide range in spatial resolution, field of view, illumination (including night imagery), and obliquity (astronauts can tilt the camera and take oblique photos).
These images have many applications in numerous fields
\cite{sprites, TLEs, astro_photos_climate_patterns, Sanchez_2022_artificial_lightning, Gaston_2022_environ_impacts_artificial_light, Small_2022_spectrometry_urban_lightscape, Schirmer_2019_nightlight_behavior}, and
critically in disaster management and response \cite{IDC_Stefanov}\footnote{\url{https://eol.jsc.nasa.gov/Collections/Disasters/ShowDisastersCollection.pl}}.

For database images we follow~\cite{Berton_2024_EarthLoc} and use a yearly-composited set of cloudless open-source satellite imagery from Sentinel-2, named \textit{S2}.
We use tiles from zoom levels 8-12 \cite{osm_zoom_levels} (expanding the range of zoom levels used in~\cite{Berton_2024_EarthLoc}), to provide a thorough train and test set that reflects the extents of the query images.
Characteristics of astronaut and satellite imagery are compared in \cref{tab:data}.

\begin{table}
\begin{center}
\begin{adjustbox}{width=\columnwidth}
\begin{tabular}{c|cc}
\toprule
Type & Queries & Database \\
\midrule
Acquisition & manual, by astronauts & automatic, by satellites \\
Number (\textit{Annotation}) & 5M (\textit{none}), 300k (\textit{weak}) & 5.2M (\textit{full}) \\
Extent &  40 to 1,357,493 sqkm &  346 to 391,776 sqkm \\
Extent percentile (5 / 95) &  239 / 39,729 sqkm &  465 / 16,026 sqkm \\
Occlusions & Yes (ISS hardware, clouds) & No \\
Obliquity \& distortion & Yes & No \\
Illumination changes & Yes (\eg day/dawn/sunset) & No \\
Years &  2000 to 2024 &  2018 to 2021 \\
Source & \url{https://eol.jsc.nasa.gov/} & \url{https://s2maps.eu} \\
Coverage & scattered, biased 
on glaciers, volcanoes etc. & dense, worldwide \\
\bottomrule
\end{tabular}
\end{adjustbox}
\end{center}
\vspace{-5mm}
\caption{\textbf{Overview of data.} Information reported about satellite imagery refers specifically to the data used in this project, not satellite imagery in general. The labeling refers to the \textit{weak} and \textit{full} labeling shown in \cref{fig:weak_full_annotation}.}
\vspace{-5mm}
\label{tab:data}
\end{table}

\subsection{Training Set}
\label{sec:train_set}
In contrast to previous work, we propose taking advantage of the 300k manually localized astronaut photos for training.
The idea is to pair these photos with matching satellite imagery from the database (\ie with enough Intersection over Union, or IoU) and apply contrastive training to train models that are robust to the domain differences between astronaut and satellite imagery.
Unfortunately, the available manual localization is only a weak annotation (see \cref{fig:weak_full_annotation}), which does not provide enough information to compute IoU. To overcome this limitation, we estimate the precise footprint for each query image as follows:
for each manually localized query, we select
as potential positives all the database images which contain the weak label point at each zoom level, and we rotate these potential positives by 90°, 180° and 270° to maximize the chance of finding a similar image among the potential positives. In practice, given five zoom levels, four rotations, and the fact that each point is covered by four images (there is overlap among the tiles), there are $5 \times 4 \times 4 = 80$ potential positives per query.
Note that different zoom levels are required because it is impossible to estimate an image extent with only a weak label, and a photo could cover an area as small as a city or as big as a continent.
We then perform image matching with SuperPoint \cite{Detone_2018_superpoint} + LightGlue \cite{Lindenberger_2023_lightglue} and the EarthMatch pipeline \cite{Berton_2024_EarthMatch} to get the footprint coordinates of each query.

This method was successful for 221k queries, with the remaining being not localizable due to either errors in the manual label, too much obstruction from clouds, or the image being of the horizon (\ie Earth limb photograph, with two corners in outer space, hence an invalid footprint).
With this now precisely localized collection of astronaut photography, we pair each query with the all database images with an IoU over $t_{iou}=0.2$ producing 865k query-database training pairs.
We then discard any pair containing a query that is included within the evaluation sets to avoid any data leakage.

\subsection{Evaluation Sets}
\label{sec:evaluation_sets}
\begin{figure}
    \begin{center}
    \includegraphics[width=0.65\columnwidth]{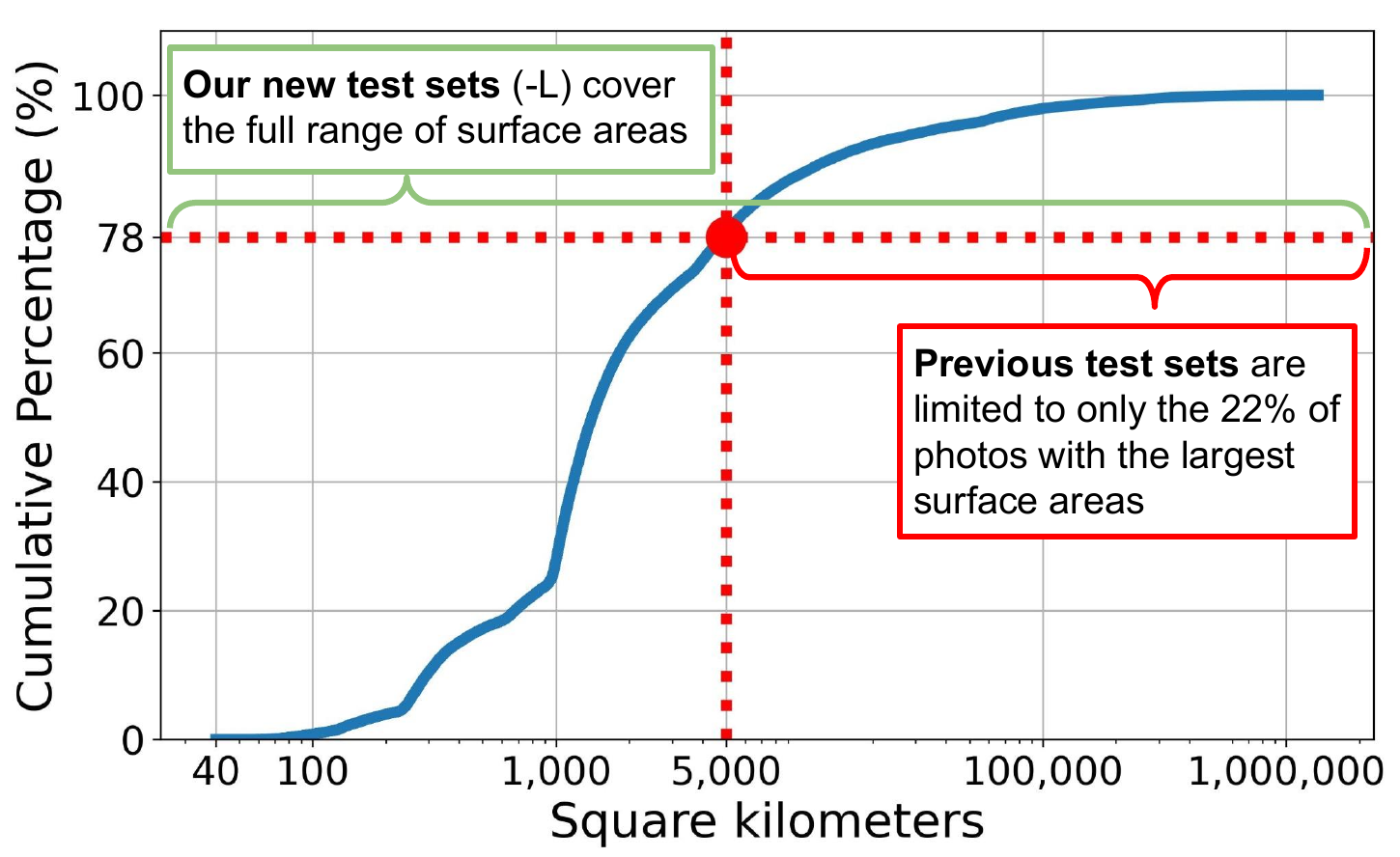}
    \end{center}
    \vspace{-5mm}
    \caption{\textbf{Distribution of queries by covered area.} The red mark at 5,000 sqkm shows that 78\% of astronaut photographs cover an area lower than 5,000 sqkm. Thus, the test sets in \cref{tab:main_table_1} do not contain this vast majority of queries, leading us to propose test sets containing all queries used for experiments in \cref{tab:main_table_2}.}
    \vspace{-5mm}
    \label{fig:sqkm_distribution}
\end{figure}

Evaluation sets are made of queries and the associated database to localize them.
Six evaluation sets were proposed in \cite{Berton_2024_EarthLoc}, covering geographical areas across five continents that mirror astronaut photography use cases like land change study and flood monitoring.
However, as detailed in their limitations \cite{Berton_2024_EarthLoc}, the test sets only contain queries whose projected area on Earth's surface is between 5,000 and 900,000 sqkm, which represents only 22\% of all astronaut photo queries, as shown in \cref{fig:sqkm_distribution}.

We therefore propose new evaluation sets which include \textit{all} available geolocated queries within an evaluation area, as well as add zoom levels 8 and 12 to the database to match the wider range of query areas.
To avoid introducing complexity, we use the same regions as the test sets from \cite{Berton_2024_EarthLoc}. These datasets were named after the geographic location of their center, like Texas, Gobi, and Amazon, so we call these new extended versions Texas-L (L for Large), Gobi-L, etc.

\section{Method}
\label{sec:method}

\subsection{Setting}
\label{sec:setting}

\begin{figure*}
    \begin{center}
    \includegraphics[width=0.9\linewidth]{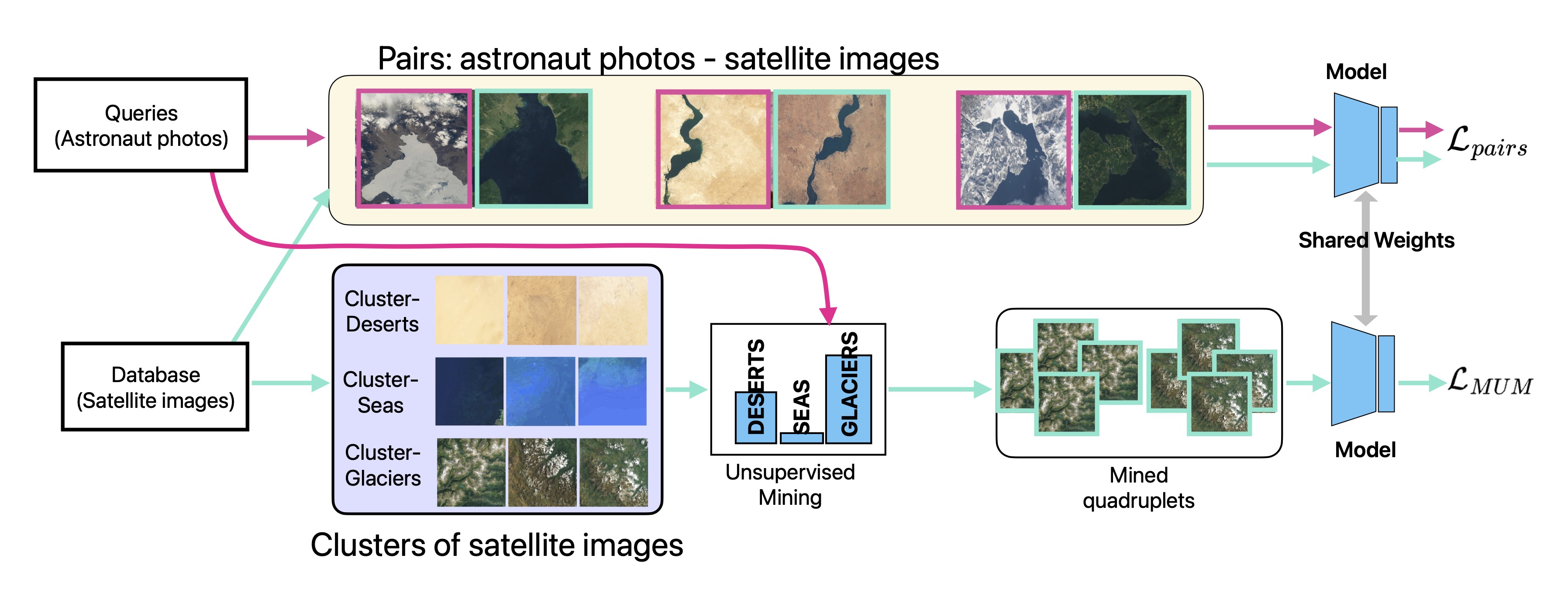}
    \end{center}
    \vspace{-8mm}
    \caption{\textbf{AstroLoc's training pipeline.}
    The upper branch feeds pairs of matching query-database images to the pairwise loss (\cref{sec:query_sat_pairwise_loss}).
    The lower branch (\cref{sec:cluster_sampling}) first creates clusters of satellite images, then queries are assigned to these clusters, which are sampled according to how many queries are assigned to each cluster (unsupervised mining).
    A training batch (mined quadruplets, \ie tuples of 4 images) is sampled from a single cluster and fed to the Multi-similarity Unsupervised Mining (MUM) loss.
    \textbf{Queries are not fed to the MUM loss, and are not used to compute the clusters} ---~they are only used to sample training data from a closer distribution to the queries'.
    }
    \vspace{-2mm}
    \label{fig:architecture}
\end{figure*}

\custompar{Task}
Astronaut Photography Localization (APL) is the task of estimating the geographic location covered by a photo (query) taken by astronauts.
APL can be approached through image retrieval: for each query, we can estimate its location by finding the most similar match in a large database of geolocated images of Earth.

\custompar{Formalization}
Throughout this paper, we rely on three sets of images:
\begin{itemize}
    \item The \textbf{database} $\mathcal{D}=\{d_1, d_2, \ldots, d_{N_{\mathcal{D}}}\}$ is a set of $N_\mathcal{D}$ \emph{satellite} images, each characterized by their RGB content and the coordinate label of their four corners, called the \textit{footprint}. Formally, the footprint for a generic image $d_i \in \mathcal{D}$ is denoted as $F_i=\{lat_1, lon_1, lat_2, lon_2, lat_3, lon_3, lat_4, lon_4\}$ where $lon_k$ and $lat_k$ denote the longitude and latitude of the $k$-th corner, respectively (\cref{fig:weak_full_annotation}).
    \item The \textbf{training queries} $\mathcal{Q}=\{q_1, q_2, \ldots, q_{N_{\mathcal{Q}}}\}$ is a set of $N_{\mathcal{Q}}$ astronaut photos,  with RGB content and footprints (localized as explained in \cref{sec:train_set}).
    \item The \textbf{test queries} $\mathcal{T}=\{t_1, t_2, \ldots, t_{N_{\mathcal{T}}}\}$ is another set of $N_{\mathcal{T}}$ astronaut photos, disjoint from $\mathcal{Q}$, each characterized by its RGB content and timestamp. The timestamp is crucial in determining the location of the ISS when the photo was taken, which enables narrowing down the search space from the entire world surface area (510M sqkm) to the local area visible from the ISS at that time (20M sqkm). Note that due to cosmic radiation bit flips, the timestamp can sometimes be incorrect: in such cases, it is necessary to conduct a worldwide search (see \cref{sec:worldwide_search}).
\end{itemize}

\custompar{Training with Astronaut Photos (Queries)}
While various approaches have been used for APL, no previous work has taken advantage of the huge dataset of 300k astronaut photos labeled by human experts.
We posit that their use at training time can lead to sizeable improvements in performance, and we present two novel techniques to use this data.
In \cref{sec:query_sat_pairwise_loss} we show how we use pairs of matching query-database images (\ie, depicting the same location) to implement a \textit{pairwise loss} (see \cref{fig:architecture}, upper branch).
Additionally, we use the photos a another time, to inform a second loss this time computed on tuples of satellite images that are sampled using the distribution of the queries (see an overview in \cref{fig:architecture} bottom). This allows us to use the entire set of database images for training (whereas the \textit{pairwise loss} can be applied only on areas covered by astronaut photos) while still leveraging the queries (see  \cref{sec:cluster_sampling}).

\subsection{Query-Satellite Pairwise Loss}
\label{sec:query_sat_pairwise_loss}

Given the cross-domain nature of the problem, we aim to take matching images from the different distributions and applying contrastive learning to encode their relationship in feature space.
To this end, we create a batch $\mathcal{P} = \{p_1, p_2, \ldots, p_{B} \}$ of query-database pairs $p_i = (q_i, d_i) \in (\mathcal{Q}, \mathcal{D})$ that have an intersection over union (IoU) higher than a threshold $t_{iou}$\footnote{For the sake of readability, we abused the notation by implying that a matching query and database images from the sets $\mathcal{Q}$ and $\mathcal{D}$ have the same index as the index of the pair they belong to.
This trick to simplify the notation is equivalent to have preliminarily sorted the sets $\mathcal{Q}$ and $\mathcal{D}$.} (see examples in \cref{fig:architecture} top).
We ensure that within a batch there are no two pairs with geographic overlap, so that for a pair $p_i\in\mathcal{P}$ all the images from other pairs $p_j \in\mathcal{P}\backslash\{p_i\}$ are negative examples.

With this setting, we define a contrastive loss that comprises two terms: an attraction term $\mathcal{L}_{pos}$ that acts between images in a matching pair, pulling their representations closer, and a repulsion term $\mathcal{L}_{neg}$, that is applied between images from different pairs (\ie, non matching), pushing them apart.
Formally, given a similarity measure $\mathcal{S}(I_1, I_2)$ between the features of two image (\eg, the cosine similarity), we define the positive loss term as
{\small
\begin{align}
        \mathcal{L}_{pos} 
        & = \frac{1}{\alpha_1 B} \sum_{i=1}^{B}
      \gamma(\alpha_1, q_i, d_i)
\end{align}
}
where $\alpha_1>0$ is a gain
and
{\small
\begin{equation}
   \gamma(x,y,z) =  \text{log} \left[ 1 + e^{-x \times \mathcal{S}(y,z)} \right]
\end{equation}
}
is the attraction function.

For the negative loss term, let us first denote with $\mathcal{P}_Q  = \{q_1, q_2, \ldots, q_B\}$ the set of all queries in the batch $\mathcal{P}$, and with $\mathcal{P}_D  = \{d_1, d_2, \ldots, d_B\}$ the set of all database images in $\mathcal{P}$. With this notation, we define $\mathcal{L}_{neg}$ as 
{\small
\begin{align}    
\begin{split}
    \mathcal{L}_{neg}
    & =  \frac{1}{\beta_1 B} \sum_{i=1}^B 
    \Big[ \varphi(\beta_1, q_i, \mathcal{P}_Q\backslash\{q_i\}) + 
     \varphi(\beta_1, q_i, \mathcal{P}_D\backslash\{d_i\}) 
    \\
    &+
     \varphi(\beta_1, d_i, \mathcal{P}_Q\backslash\{q_i\})
 + 
     \varphi(\beta_1, d_i, \mathcal{P}_D\backslash\{d_i\})
    \Big]
    \end{split}
\end{align}
}

where $\beta_1>0$ is a gain and
{\small
\begin{equation}
    \varphi(x,y,\mathcal{Z}) = \text{log} \left( 1 + \sum_{j=1}^{|\mathcal{Z}|} e^{x \times \mathcal{S}(y,\mathcal{Z}_j)} \right)
\end{equation}
}
is the repulsion function.
The overall loss is 
{\small
\begin{equation}
    \mathcal{L}_{pairs} = \mathcal{L}_{pos} + \mathcal{L}_{neg}
\end{equation}
}

\subsection{Unsupervised Mining}
\label{sec:cluster_sampling}

\begin{figure*}
    \begin{center}
    \includegraphics[width=0.99\linewidth]{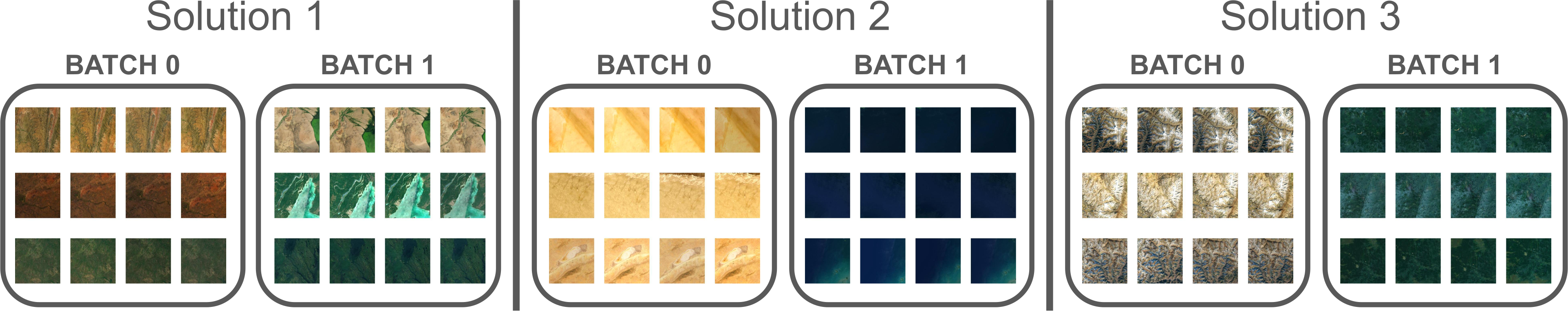}
    \end{center}
    \vspace{-5mm}
    \caption{\textbf{Examples of training batches, using the three different sampling solutions presented.} For each solution, we show two examples of batches with batch size 12 (\ie 3 quadruplets), so that each image has 3 positives and 8 negatives.
    \textbf{Solution 1} leads to training a non-robust model, because of the lack of hard negatives, as quadruplets are randomly sampled and are often very diverse between each other.
    \textbf{Solution 2} provides hard negatives, but it creates batches with uninformative and feature-less images (\eg \textit{batch 1} is from a cluster of images of seas next to coasts) which hurt the training process.
    \textbf{Solution 3} allows to simultaneously provide hard negatives, discard uninformative batches, and focus on the most salient imagery, by sampling images from clusters that reflect the queries' distribution.
    }
    \vspace{-1mm}
    \label{fig:batch_examples}
\end{figure*}

Although the use of the Query-Satellite Pairwise Loss by itself leads to state-of-the-art results (see \cref{sec:experiments}), it does not fully exploit the large quantity of available satellite imagery worldwide, as it only uses images that overlap with astronaut photographs.
Therefore, we propose to use a second loss: in the next paragraphs, we explain the design of this innovative technique, starting from a naive implementation and developing into our full-fledged MUM loss.

\custompar{Solution 1: naive sampling}
The simplest approach to train a retrieval model on large volumes of satellite imagery would be to directly apply a contrastive loss.
Common training pipelines rely on using quadruplets \cite{Musgrave_2020_ML_reality, Alibey_2022_gsvcities} (\ie 4 images) from each class, and stacking multiple quadruplets within a batch to ensure that each image has exactly 3 positives (from the same class) and a multitude of negatives.
However, this naive approach would not train a robust model, because random sampling of classes would lead to very few hard negatives (see \cref{fig:batch_examples} left), which are crucial to train robust image retrieval models.

\custompar{Solution 2: database clustering}
To improve over naive sampling we could first compute features for all images within our database, and then cluster them with a k-means algorithm in feature space.
This leads to the creation of $K$ clusters that share similar visual characteristics: for example one cluster containing images of forests, another one with images of deserts, and so on.
These clusters can then be used to create training batches: for example there would be one batch of forest images, one batch of desert images, et cetera, leading to harder negatives within batches.
While this solution leads to better results (see \cite{Berton_2024_EarthLoc}), one issue would arise: some batches may contain only uninformative images, like feature-less images of deserts or seas, (see \cref{fig:batch_examples} center), which can hurt the learning process.

\custompar{Solution 3: unsupervised mining}
We finally propose to not only create clusters from database images, but also to sample them according to the distribution of the astronaut images:
intuitively this means that, as astronauts take very few photographs of deserts, and many photographs of glaciers, the clusters of desert-images will be sampled seldom, while the clusters with glaciers will be sampled more often (see \cref{fig:batch_examples} right).

To formalize this concept, we split the database images into a set of $K$ clusters $\mathcal{C}_1, \ldots, \mathcal{C}_K$ using a standard k-means algorithm in feature space, such that
{\small
\begin{equation}
    \bigcup_{k=1}^{K} \mathcal{C}_k = \mathcal{D}
\end{equation}
}
Secondly, we compute features for the training queries (\ie astronaut photos), and assign each to one of the K clusters $C$: we then denote with $b_1, \ldots, b_K$ the size of the bins derived from assigning the features of the training queries to the $K$ clusters.
Clusters are then sampled according to
{\small
\begin{equation}
    k\sim B(Q, 1, k)
    \label{eq:weighted_sampling2}
\end{equation}
}
where $B$ denotes the weighted distribution according to the bins $b_1, \ldots, b_K$, \ie such that 
{\small
\begin{equation}
  Pr(k) = \frac{b_k}{\sum_{i=1}^{K}{b_i}}
\end{equation}
}
meaning that clusters with more queries are sampled more often for the creation of our training batches.
We emphasize that the training batches are made only of database images, and that the queries are only used to calculate how often each cluster should be sampled, as shown in \cref{fig:architecture}.
Finally, given a series of quadruplets from cluster $\mathcal{C}_K$ (sampled according to \cref{eq:weighted_sampling2}), we plug them into a contrastive loss (in our case a Multi-Similarity loss \cite{Wang_2019_multi_similarity_loss}). Therefore, the second loss in our pipeline is
{\small
\begin{equation}
    \begin{split}
        \mathcal{L}_{MUM} = \frac{1}{4H}\sum_{i=1}^{H} \left[\frac{1}{\alpha_2} 
        \sum_{j=1}^4 \text{log} 
        \left( 1+ \sum_{d\in h_i\backslash\{h_{i_{j}}\}}e^{-\alpha_2\mathcal{S}(h_{{i}_j}, d)} \right) \right. \\ 
        + \left. \frac{1}{\beta_2} \sum_{j=1}^4 \text{log} \left( 1+ \sum_{d\in\mathcal{H}_k\backslash \{h_i\}}e^{+\beta_2 \mathcal{S}(h_{{i}_j}, d)} \right) \right]
    \end{split}
\end{equation}
}
where $\alpha_2$ and $\beta_2$ are positive hyperparameters, and MUM stands for \textbf{M}ulti-similarity with \textbf{U}nsupervised \textbf{M}ining.

Among the large number of mining techniques in literature, our Unsupervised Mining stands out in two ways:
\begin{enumerate}
  \item it is the only mining that uses one distribution (\ie $Q$) to learn how to sample from another distribution (\ie $\mathcal{D}$);
  \item unlike common mining techniques used in localization and place recognition pipelines \cite{Arandjelovic_2018_netvlad, AliBey_2022_BMVC, Musgrave_2020_PyTorchML, Ge_2020_sfrs}, our miner does not require labels of queries, so we could potentially use all of the 5M existing astronaut photos (even the unlabeled ones). In practice we use only training queries to ensure avoiding any kind of test data leakage.
\end{enumerate}
Finally, our final loss is the sum of the two losses:{\small
\begin{equation}
    \mathcal{L} = \lambda_{1} \mathcal{L}_{pairs} + \lambda_{2} \mathcal{L}_{MUM}
\end{equation}
}

\begin{table*}
\begin{center}
\begin{adjustbox}{width=0.999\textwidth}
\begin{tabular}{c|ccc c ccc c ccc c ccc c ccc c ccc}
\toprule
\multirow{2}{*}{Method} &
\multicolumn{3}{c}{Texas (6k / 34k)} &&
\multicolumn{3}{c}{Alps (2k / 53k)} &&
\multicolumn{3}{c}{California (4k / 30k)} &&
\multicolumn{3}{c}{Gobi Desert (1k / 54k)} &&
\multicolumn{3}{c}{Amazon (1k / 19k)} &&
\multicolumn{3}{c}{Toshka Lakes (2k / 63k)} \\
\cline{2-4} \cline{6-8} \cline{10-12} \cline{14-16} \cline{18-20} \cline{22-24}
& R@1 & R@10 & R@100 && R@1 & R@10 & R@100 && R@1 & R@10 & R@100 && R@1 & @10 & R@100 && R@1 & R@10 & R@100 && R@1 & R@10 & R@100 \\
\midrule
Nadir         & 2.4 & - & - && 1.2 & - & - && 2.4 & - & - && 1.8 & - & - && 3.1 & - & - && 1.4 & - & - \\
Random Choice &  0.2 &  1.7 & 15.5 &&  0.1 &  1.1 & 11.6 &&  0.2 &  2.3 & 20.1 &&  0.1 &  1.0 & 13.2 &&  0.1 &  1.1 & 11.5 &&  0.2 &  1.2 &  9.1 \\
\midrule
TorchGeo \cite{Stewart_2022_TorchGeo} (SSL w SeCo \cite{Yao_2021_seco}) &  6.1 & 15.6 & 41.7 &&  7.4 & 20.2 & 49.2 &&  5.1 & 14.5 & 37.1 &&  3.7 & 14.0 & 38.9 &&  4.6 & 13.3 & 32.9 &&  5.7 & 15.6 & 38.5 \\
TorchGeo \cite{Stewart_2022_TorchGeo} (SSL w GASSL \cite{Ayush_2021_geography}) &  9.7 & 22.8 & 46.4 &&  9.1 & 23.1 & 50.5 && 13.3 & 31.4 & 58.8 &&  6.3 & 17.5 & 45.4 && 8.3 & 20.3 & 40.1 && 20.4 & 38.6 & 64.2 \\
OGCL UAV-View \cite{Deuser_2023_ogcl_uav_view}  \cite{Dosovitskiy_2021_vit} & 17.6 & 33.2 & 55.9 && 14.6 & 33.2 & 63.9 && 22.8 & 48.1 & 74.9 &&  7.6 & 22.8 & 50.1 && 20.4 & 39.1 & 62.5 && 31.8 & 51.8 & 74.4 \\
MBEG \cite{Zhu_2023_uav_backbone_winnerUAV_mbeg}  &  7.0 & 17.6 & 35.1 &&  6.6 & 19.3 & 45.9 &&  8.7 & 20.7 & 41.5 &&  4.4 & 15.0 & 38.0 &&  6.4 & 17.1 & 39.3 &&  8.1 & 20.7 & 49.1 \\
AnyLoc \cite{Keetha_2023_AnyLoc} & 44.1 & 68.7 & 87.8 && 40.7 & 70.8 & 92.0 && 48.7 & 75.0 & 91.6 && 28.7 & 57.0 & 81.7 && 38.6 & 63.8 & 86.2 && 63.7 & 84.5 & 96.3 \\
EarthLoc \cite{Berton_2024_EarthLoc}         & 55.9 & 73.0 & 88.3 && 58.4 & 76.8 & 89.5 && 58.0 & 76.0 & 91.4 && 51.1 & 67.5 & 86.5 && 47.2 & 67.9 & 84.6 && 72.2 & 85.0 & 93.3 \\
EarthLoc++ \cite{Berton_2024_EarthLoc} & \underline{80.0} & \underline{89.4} & \underline{95.9} & & \underline{80.6} & \underline{91.2} & \underline{96.7} & & \underline{82.9} & \underline{92.0} & \underline{97.9} & & \underline{67.6} & \underline{84.6} & \underline{94.6} & & \underline{73.6} & \underline{85.5} & \underline{93.1} & & \underline{90.1} & \underline{95.3} & \underline{98.2} \\
AstroLoc  & \textbf{96.1} & \textbf{98.7} & \textbf{99.7} & & \textbf{98.1} & \textbf{99.5} & \textbf{99.8} & & \textbf{97.4} & \textbf{99.2} & \textbf{99.8} & & \textbf{94.6} & \textbf{99.2} & \textbf{99.9} & & \textbf{93.0} & \textbf{96.9} & \textbf{99.1} & & \textbf{99.0} & \textbf{99.6} & \textbf{99.9} \\
\bottomrule
\end{tabular}
\end{adjustbox}
\end{center}
\vspace{-2mm}
\caption{
\textbf{Recalls on APL test sets from EarthLoc \cite{Berton_2024_EarthLoc},} from which we source most of the numbers in the table. 
AstroLoc comfortably and consistently achieves recall over 90\%, outperforming all other methods by a large margin.
Numbers next to each test set name (\eg Texas 6k / 34k) represent number of queries / database images. R@N indicates the recall-at-N. Best results \textbf{bold}, second best \underline{underlined}.
}
\vspace{-1mm}
\label{tab:main_table_1}
\end{table*}

\begin{table*}
\begin{center}
\begin{adjustbox}{width=0.999\textwidth}
\begin{tabular}{ccccc|ccc ccccccccccccccc}
\toprule
\multirow{2}{*}{Method} & Features & Memory & Total latency & Per query &
\multicolumn{2}{c}{Texas-L (21k / 179k)} &&
\multicolumn{2}{c}{Alps-L (10k / 261k)} &&
\multicolumn{2}{c}{California-L (12k / 166k)} &&
\multicolumn{2}{c}{Gobi Desert-L (2k / 284k)} &&
\multicolumn{2}{c}{Amazon-L (3k / 101k)} &&
\multicolumn{2}{c}{Toshka Lakes-L (9k / 299k)} \\
\cline{6-7} \cline{9-10} \cline{12-13} \cline{15-16} \cline{18-19} \cline{21-22}
& dimension & (GB) &  (hh:mm) & latency (ms)
& R@1 & @100 && R@1 & @100 && R@1 & @100 && R@1 & @100 && R@1 & @100 && R@1 & @100 \\
\midrule
AnyLoc  & 49152 & 235 GB & 23:51 & 291 & 29.2 & 72.2 & & 30.9 & 78.9 & & 31.7 & 76.3 & & 22.9 & 68.2 & & 20.4 & 63.5 & & 34.9 & 76.8 \\
EarthLoc & 4096 &  19 GB &  1:12 &  21 & 44.0 & 75.0 & & 52.9 & 82.6 & & 49.1 & 80.4 & & 33.9 & 70.1 & & 34.9 & 70.6 & & 63.3 & 86.5 \\
EarthLoc++&2048 &   9 GB &  1:42 &  12 & 72.4 & 91.5 & & 79.4 & 94.7 & & 75.4 & 92.9 & & 62.6 & 87.2 & & 62.6 & 89.2 & & 83.4 & 95.1 \\
AstroLoc & 2048 &   9 GB &  1:42 &  12 & \textbf{91.1} & \textbf{98.5} & & \textbf{94.6} & \textbf{99.2} & & \textbf{92.1} & \textbf{98.7} & & \textbf{84.2} & \textbf{96.2} & & \textbf{83.8} & \textbf{96.8} & & \textbf{94.8} & \textbf{99.0} \\
\bottomrule
\end{tabular}
\end{adjustbox}
\end{center}
\vspace{-5mm}
\caption{\textbf{Results on our extended versions of the test sets from \cref{tab:main_table_1}.}
``-L" (for ``large") test sets better represent the performance that an APL model would achieve if deployed,  as no filtering has been applied on test queries. The two numbers (\eg Texas 21k / 179k) represent the number of queries and database images.
\textit{Memory}, \textit{Total latency} and \textit{Per query latency} refer to testing on the Texas dataset. \textit{Memory} is the memory required to store features; \textit{Total latency} is the time to run the test (including database feature extraction); and \textit{Per query latency} is the time needed to localize one query, which consists of query feature extraction and kNN search (kNN is the primary bottleneck). All tests are performed with an A100 and a 32-core CPU, using FAISS \cite{Douze_2024_faiss} for efficient kNN.
}
\label{tab:main_table_2}
\end{table*}

\section{Experiments}
\label{sec:experiments}

\subsection{Implementation Details}
\label{sec:Implementation_details}

\custompar{Training}
We set the hyperparameters as follows: $t_{iou}=0.2, \alpha_1=1, \alpha_2=1, \beta_1=50, \beta_2=50, \lambda_1=1, \lambda_2=1, K=50$, batch size $=$ 48 (48 pairs for pair loss, 48 quadruplets for MUM loss), learning rate $=$ 5e-5, Adam optimizer.
We use as architecture a DINO-v2-base backbone \cite{Oquab_2023_dinov2} with SALAD \cite{Izquierdo_2024_SALAD} and a linear layer to reduce feature dimensionality (from 8448 to 2048) (\ie a similar model to AnyLoc \cite{Keetha_2023_AnyLoc} while being over 10 times lighter than AnyLoc, which uses DINO-v2-giant).
Training runs for 30k iterations.
As in \cite{Berton_2024_EarthLoc}, the Texas dataset is used as validation.
The features for our Unsupervised Mining are computed periodically (every 5000 iterations) while the model is undergoing training, as standard practice in virtually every mining procedure \cite{Arandjelovic_2018_netvlad, AliBey_2022_BMVC, Berton_2024_EarthLoc, Musgrave_2020_PyTorchML}. With this setup, mining increases the training time by less than 10\%.

\custompar{Evaluation}
We follow the protocol from \cite{Berton_2024_EarthLoc} by doing image retrieval with an augmented dataset (\ie applying four 90° rotations to each image),
and using as metric the recall@N, as the percentage of queries where at least one of the top-N predictions is a correct match to the query.

\subsection{Results}
In \cref{tab:main_table_1} we report experiments on the test sets proposed by \cite{Berton_2024_EarthLoc}.
For fairness, we include results with a more powerful version of EarthLoc that uses the same architecture and training data of AstroLoc, which we refer to as EarthLoc++.

Results clearly show that, whereas previously there was no dominant method between EarthLoc and AnyLoc, the introduction of AstroLoc provides a new state-of-the-art model which significantly outperforms all previous, achieving a near perfect (above 99\%) recall@100 on all six evaluation sets.
Note that the recall@100 is more relevant in the real world than recall@1, since it is common practice to re-rank the top-N predictions with image matching methods \cite{Berton_2024_EarthMatch, Stoken_2023_CVPR, Trivigno_2024_CVPR}.
Some qualitative results of queries and their predictions are in \cref{fig:teaser}, with many more in the supplementary.

\custompar{New test sets}
Here we compute experiments on the newly extended test sets described in \cref{sec:evaluation_sets}, which provide a setting that is more similar to the real world scenario and more challenging. Results, reported in \cref{tab:main_table_2}, illustrate that AstroLoc is able to outperform previous models by an even wider margin, and presents recall@100 consistently above 96\% even in these more comprehensive and complex cases.

\subsection{Other Space to Ground Use Cases}
In this paper we test the robustness of AstroLoc by performing experiments on two related tasks: the lost-in-space problem and historical space imagery localization.
Given the lack of space, we present a thorough explanation of the tasks, motivations, experimental details and results in the supplementary, while only offering a brief summary here.
\label{sec:lost_in_space}

\begin{table}
\begin{center}
\begin{adjustbox}{width=0.9\columnwidth}
\begin{tabular}{c|c|ccccc}
\toprule
Method & \# Params (M) & R@1 & R@5 & R@10 & R@20 & R@100 \\
\midrule
AnyLoc        & 1136 &4.4 &  9.6 & 13.4 & 19.0 & 40.0 \\
EarthLoc      & \underline{27.6} &3.3 &  7.0 &  8.8 & 12.0 & 27.6 \\
AstroLoc      & 105 &\textbf{52.7} & \textbf{63.9} & \textbf{68.7} & \textbf{73.3} & \textbf{83.7}\\
AstroLoc-tiny & \textbf{27.2} &\underline{36.7} & \underline{49.4} & \underline{55.3} & \underline{61.6} & \underline{74.5} \\
\bottomrule
\end{tabular}
\end{adjustbox}
\end{center}
\vspace{-5mm}
\caption{\textbf{Results of Lost-in-Space satellite problem.} Using the dataset from VINSat \cite{McCleary_2024_vinsat} as queries, AstroLoc achieves good results on this additional task. AstroLoc-tiny is a tiny version more suitable for deployment on a nanosatellite, with 27M model parameters and 512 dimensional features (see supplementary).}
\label{tab:vinsat}
\end{table}

\custompar{The Lost-in-Space satellite problem} has the goal of identifying the location/orbit of nanosatellites through computer vision \cite{McCleary_2024_vinsat}, requiring photos be searched for over the entire planet.
We use as queries the images from McLeary et al. \cite{McCleary_2024_vinsat}, and a worldwide database at zoom level 9 (12k images).
Results in \cref{tab:vinsat} highlight the robustness and adaptability of AstroLoc, which is the only method to achieve a R@1 over 50\% on the retrieval-formulation of lost in space, outperforming all other methods by at least 45\% of R@1.

\begin{table}
\begin{center}
\begin{adjustbox}{width=0.8\columnwidth}
\begin{tabular}{c|ccccc}
\toprule
Method & R@1 & R@5 & R@10 & R@20 & R@100 \\
\midrule
AnyLoc   & 19.6 & 32.7 & 40.5 & 47.2 & \underline{65.2} \\
EarthLoc & \underline{29.5} & \underline{40.1} & \underline{43.5} & \underline{48.7} & 64.5 \\
AstroLoc & \textbf{82.0} & \textbf{89.9} & \textbf{91.9} & \textbf{94.2} & \textbf{96.7} \\
\bottomrule
\end{tabular}
\end{adjustbox}
\end{center}
\vspace{-5mm}
\caption{\textbf{Results on historical imagery localization.} The 704 queries are 40 year-old photos from the early days of the Space Shuttle. Searched is conducted across the whole globe.}
\vspace{-2mm}
\label{tab:sshuttle}
\end{table}

\custompar{Historical space imagery localization}
Historical imagery of Earth from space is a unique source of data to understand how Earth has changed over decadal time spans.
We perform experiments on 704 queries from early Space Shuttle days (1981-1984), on a worldwide database, and report results in \cref{tab:sshuttle}.
We empirically find that AstroLoc achieves strong results even on these old film photographs, showcasing its robustness to various types of domain changes.

\subsection{Ablations}
\label{sec:ablations}
We compute ablations on the losses and mining in \cref{tab:ablation}, where results clearly show the strong impact from each module in our pipeline.
Most importantly, the combination of these components presents the optimal characteristic of orthogonality, in that their mixture has notably higher results than each of the singular elements.

\begin{table}
\begin{center}
\begin{adjustbox}{width=0.999\columnwidth}
\begin{tabular}{cccc|ccc c ccc c ccc c ccc c ccc c ccc}
\toprule
Pair Loss &
Sol. 1 &
Sol. 2 &
Sol. 3 &
\multicolumn{3}{c}{Texas-L} &&
\multicolumn{3}{c}{Alps-L}
\\
\cline{5-7} \cline{9-11}
(Sec 4.2) &
(Sec 4.3) &
(Sec 4.3) &
(Sec 4.3)
& R@1 & R@10 & R@100 &
& R@1 & R@10 & R@100 \\
\midrule
\checkmark &&&& 
83.6 & 93.1 & 97.5 & & 87.2 & 95.0 & 98.2 \\
& \checkmark &&& 
67.6 & 79.2 & 89.3 & & 76.5 & 87.1 & 93.9 \\
&& \checkmark && 
72.4 & 83.3 & 91.5 & & 79.4 & 89.1 & 94.7 \\
&&& \checkmark & 
82.2 & 91.1 & 97.1 & & 86.9 & 94.5 & 97.9 \\
\midrule
\checkmark & \checkmark &&&
84.0 & 93.3 & 97.5 & & 87.9 & 95.1 & 98.0 \\
\checkmark && \checkmark && 
86.5 & 94.1 & 97.7 & & 91.5 & 96.5 & 98.6 \\
\checkmark &&& \checkmark & 
\textbf{91.1} & \textbf{96.2} & \textbf{98.5} & & \textbf{94.6} & \textbf{97.8} & \textbf{99.2} \\
\bottomrule
\end{tabular}
\end{adjustbox}
\end{center}
\vspace{-5mm}
\caption{\textbf{Ablation over components and variation of the loss.}
}
\label{tab:ablation}
\end{table}

\subsection{Toward Worldwide Search}
\label{sec:worldwide_search}
Astronaut photographs are associated with a timestamp which, combined with the known orbit of the ISS, can be used to obtain the position of the camera when the photo was taken, allowing the search space to be reduced from the entire extent of the Earth to just 20M sqkm (the area of Earth's surface visible from the ISS).
However, due to high-energy cosmic particles causing bit-flips in cameras, the timestamp can be unreliable, leaving no information about which area of the Earth the photo is taken near.

We therefore propose the task of worldwide astronaut photography localization, where the database covers the entire world with 881k images.
Results (\cref{tab:worldwide}) show that AstroLoc achieves a recall@100 of 96.8\%, proving its robustness even in this highly challenging scenario.

\begin{table}
\begin{center}
\begin{adjustbox}{width=0.95\columnwidth}
\begin{tabular}{c|ccccc}
\toprule
Method & R@1 & R@5 & R@10 & R@20 & R@100 \\
\midrule
EarthLoc  & 40.3 & 49.5 & 53.2 & 57.1 & 66.6 \\
AstroLoc  & 86.9 & 91.9 & 93.4 & 94.6 & 96.8 \\
\bottomrule
\end{tabular}
\end{adjustbox}
\end{center}
\vspace{-3mm}
\caption{\textbf{World-wide search.} Using the same query sets as \cref{tab:main_table_2}, but with a database encompassing the entire Earth. Results computed for the best performing methods only, except AnyLoc \cite{Keetha_2023_AnyLoc}, due to its large features (49,152-D vs EarthLoc's 4096-D) requiring \>600GB of RAM to store features to compute kNN.}
\label{tab:worldwide}
\end{table}

\section{Conclusions}
\label{sec:conclusions}
We tackle the task of Astronaut Photography Localization (APL),
and through newly computed labels and ad hoc training techniques, obtain
AstroLoc, which shows impressive results on all APL datasets.
Furthermore, we find that AstroLoc can be utilized for other real-world applications, showing robustness to various domains.
Finally, we note that we have already used AstroLoc (in conjunction with EarthMatch's post-processing) to provide localization for a staggering 500k photos, and anticipate that in a few months the backlog of non-localized photos will be nearly empty for the first time since the launch of the ISS.

\paragraph{Acknowledgements.}
\small{We acknowledge the Cineca award under the Iscra initiative, for the availability of high performance computing resources.
This work was supported by CINI.
Project supported by ESA Network of Resources Initiative.
This study was carried out within the project FAIR - Future Artificial Intelligence Research - and received funding from the European Union Next-GenerationEU (Piano nazionale di ripresa e resilienza (PNRR) – missione 4 componente 2, investimento 1.3 – D.D. 1555 11/10/2022, PE00000013). This manuscript reflects only the authors’ views and opinions, neither the European Union nor the European Commission can be considered responsible for them.
European Lighthouse on Secure and Safe AI – ELSA, Horizon EU Grant ID: 101070617
}

\clearpage
\setcounter{page}{1}
\maketitlesupplementary

\section{Supplementary}
In this Supplementary we provide additional information on the Lost-in-Space satellite problem (\cref{sec:supp_lost_in_space}),
the Historical space imagery localization problem (\cref{fig:supp_sshuttle}), and finally we showcase a large number of qualitative results \cref{sec:supp_qualitatives}.

\subsection{Lost-in-Space satellite problem}
\label{sec:supp_lost_in_space}

\custompar{Overview} The goal of ``lost in space" is to identify the location/orbit of nanosatellites using computer vision. Existing solutions either involve bulky and expensive ($\sim$10k\$) GPS receivers, months-long tracking via radio communications through ground stations \cite{Skinner_2022_cubesat_radar} (a considerable fraction of CubeSats remain
unidentified more than 250 days after launch \cite{McCleary_2024_vinsat}), or use the recent VINSat \cite{McCleary_2024_vinsat}, a computer vision solution that localizes the satellite only if/when it flies above a set of predefined landmarks.
With our more general approach to localization, we can instead localize the queries (collected in real time by the satellite) anywhere on Earth, without being constrained to a set of predefined locations or landmarks.

We apply AstroLoc by reformulating the task as image retrieval, using the dataset from McCleary et al. \cite{McCleary_2024_vinsat}. This dataset contains Sentinel-2 mosaics, which are not cloudless or seamlessly processed, leading to sharp boundaries and occasional oblique views (see \cref{fig:supp_vinsat}, row 1 col 2). 
Though Sentinel 2 captures images nadir-facing (straight down), some images in this set (\cref{fig:supp_vinsat}, row 1 col 2) appear to have been transformed to mimic oblique views. These conditions increase the challenge of this dataset. All images cover an area between 35k and 55k sqkm, meaning that our database can be built with only satellite images of zoom level 9, reducing the number of images required for worldwide coverage to just 12k.
We therefore construct an image retrieval task with all 2500 images from \cite{McCleary_2024_vinsat} as queries, and 12k database images.

\custompar{Results} Tab. 4 (of the main paper) shows AstroLoc's superior performance in an out-of-distribution task, outperforming AnyLoc and EarthLoc by $\sim$50 R@1. These gains highlight the impact of AstroLoc's training setup improvements. Additionally, even the smaller AstroLoc-tiny model outperforms AnyLoc with only 2\% of the parameters.

\custompar{Embedded Use} To better understand if this model could be actually used on a nanosatellite, we also provide results with a tiny version of AstroLoc, based on the smallest version of DINO-v2 and with output dimension of 512.
This version achieves 36.7\% R@1 and 74.5\% R@100 with only 27M parameters (25\% of the full AstroLoc), making it lightweight enough to fit on an embedded system in a nanosatellite.
The memory required to store database features with AstroLoc-tiny is 
$12k \times 512 \times 4 \times 4 = 98 \text{ MB}$: 12k images with 512 dimensional features, each repeated 4 times due to the test time rotation augmentation and each element taking 4 bytes due to float32 encoding.
Further memory reduction can be achieved through compression with methods like product quantization \cite{Jegou_2011_productQ} for the features or by quantizing or pruning the model itself.
Given the smaller memory footprint of this model and database, it is feasible to use AstroLoc to run real time localization of nanosatellites. Combined with the state estimation techniques from ~\cite{McCleary_2024_vinsat}, AstroLoc can power a fast, accurate, and low cost orbit determination solution.

\subsection{Historical space imagery localization}

Historical imagery of Earth from space represents a unique source of data to understand how Earth has changed over decadal time spans.
Similar to photographs taken by astronauts aboard the ISS, lots of historical imagery from space, taken from the Space Shuttle, lacks localization information.
Although efforts at manually localizing these photos have been made, we note that (1) a large number of these images still lack location information, and (2) these images are only weakly labeled with a single location of any pixel within the image, which is often noisy and does not represent its full extent.

Given these shortcomings, we seek to understand the performance of APL systems in localizing early photographs (1981-1984) from the Space Shuttle, which were originally taken with film cameras and then digitized. As such, they have different photometric characteristics from more recent photography, and are known\footnote{\url{https://eol.jsc.nasa.gov/FAQ/\#photoQuality}} to often require color correction (note the blue hue in many images in \cref{fig:supp_sshuttle}). We first precisely localize 704 images with the pipeline described in Sec. 3.1, and then compute localization results, reported in Tab. 5.
We empirically find that AstroLoc achieves strong results even on these old photographs, showcasing its robustness to various types of domain changes. Comparing against performance on APL evaluation sets (Table 2 and Table 3 of the main paper), AstroLoc has the smallest drop in recall (at all N) on the out-of-distribution historical imagery out of all methods tested.

\begin{figure*}[h]
    \begin{center}
    \includegraphics[width=0.99\linewidth]{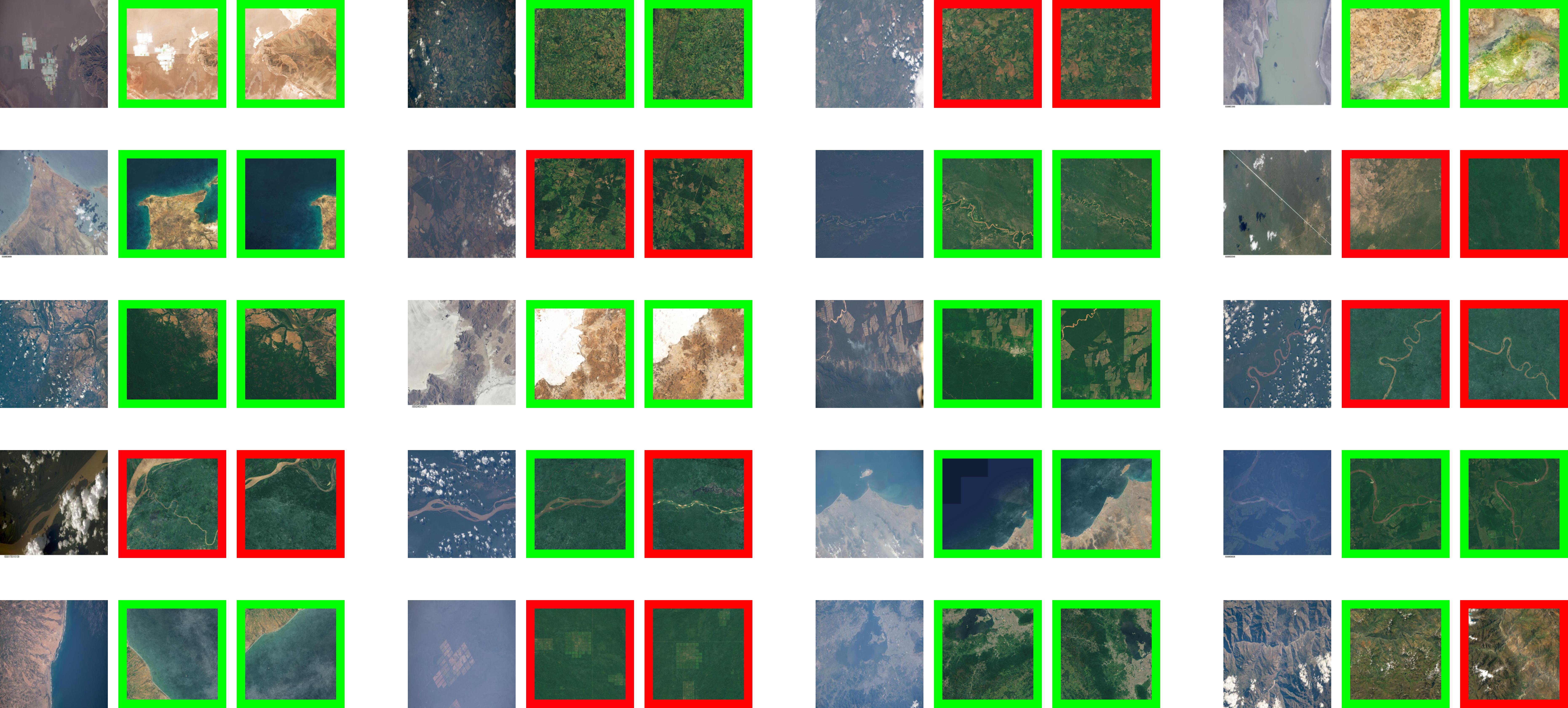}
    \end{center}
    \caption{\textbf{Qualitative examples from the Amazon-L test set.} Each triplet shows one query and its top-2 predictions, red if wrong and green if correct. }
    \label{fig:supp_astroloc_Amazon}
\end{figure*}

\subsection{Qualitative Results}
\label{sec:supp_qualitatives}
\begin{figure*}
    \begin{center}
    \includegraphics[width=0.999\linewidth]{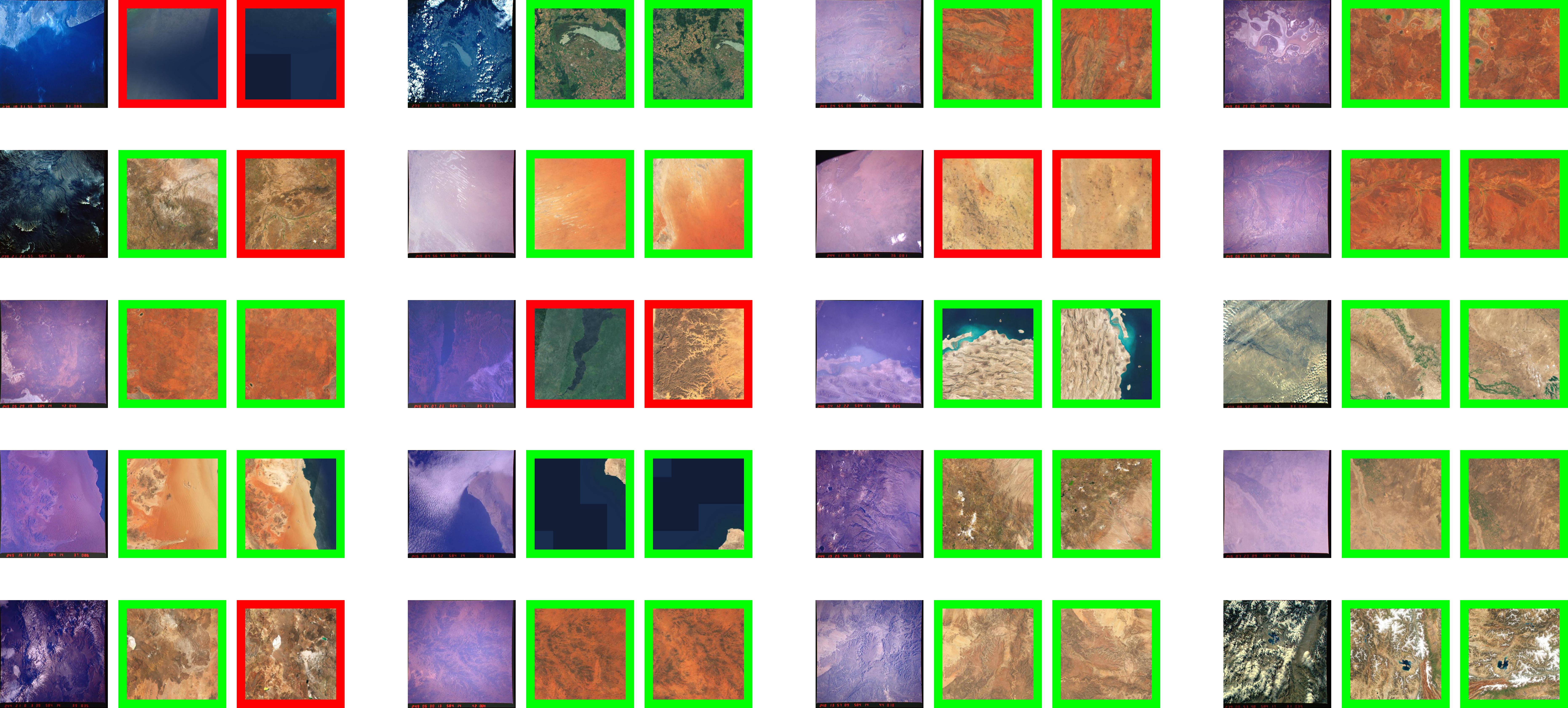}
    \end{center}
    \caption{\textbf{Qualitative examples from the historical Space Shuttle imagery.} Each triplet shows one query and its top-2 predictions, red if wrong and green if correct. The queries were taken with analog cameras between 1981 and 1984 and then later digitized.}
    \label{fig:supp_sshuttle}
\end{figure*}

\begin{figure*}
    \begin{center}
    \includegraphics[width=0.999\linewidth]{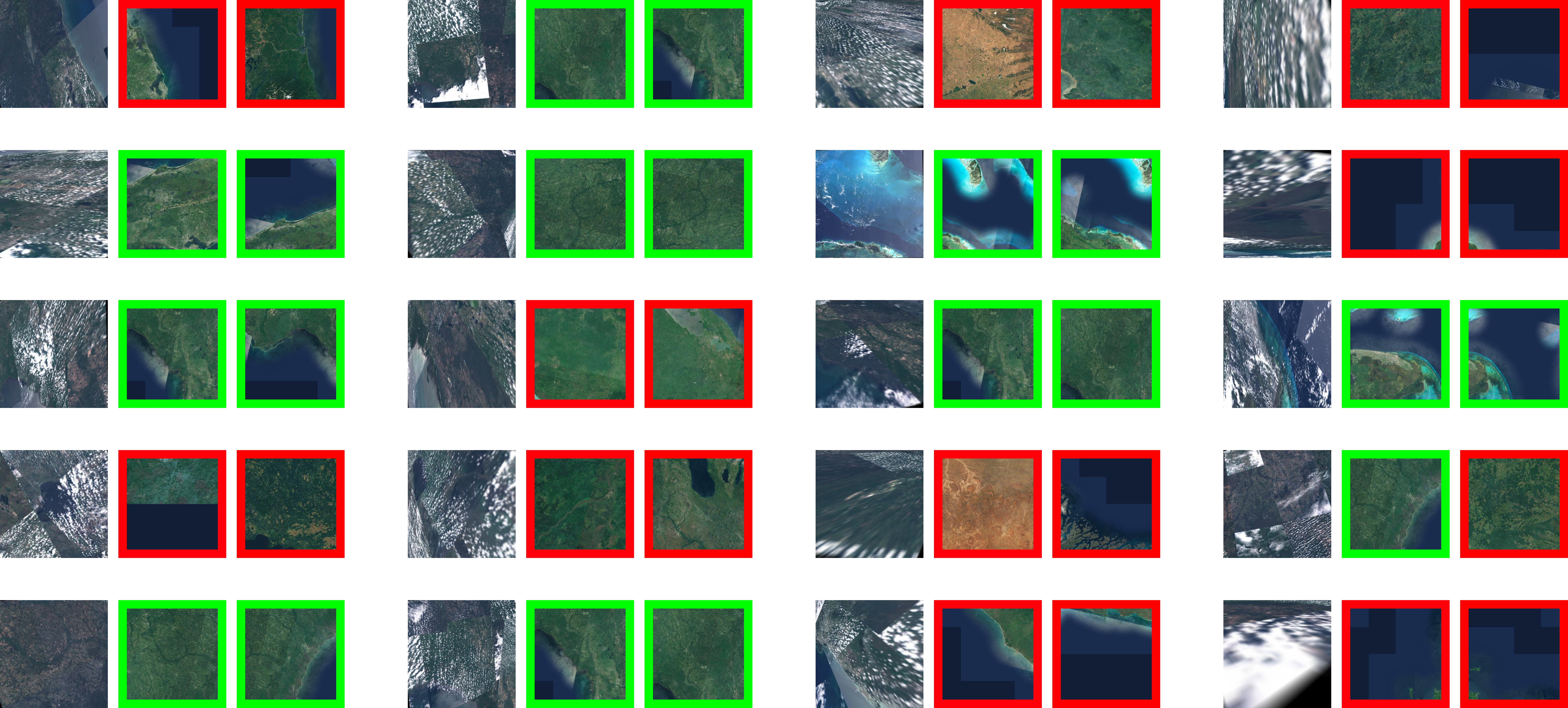}
    \end{center}
    \caption{\textbf{Qualitative examples from the VINSat dataset \cite{McCleary_2024_vinsat}.} Each triplet shows one query and its top-2 predictions, red if wrong and green if correct. Queries are mosaics of Sentinel 2 imagery.}
    \label{fig:supp_vinsat}
\end{figure*}

\cref{fig:supp_astroloc_Amazon}, \ref{fig:supp_sshuttle}, \ref{fig:supp_vinsat} show qualitative AstroLoc results on astronaut photos, historical Earth from space imagery from the Space Shuttle, and satellite imagery from the McCleary\cite{McCleary_2024_vinsat} dataset, respectively. There are four examples in each row, with each example showing the query photo (leftmost) and the top-2 retrieval results (\ie most similar, as ranked by AstroLoc). Correct predictions, defined as those that have \textit{any} overlap with the query, are outlined in green, and incorrect predictions are outlined in red. 
\custompar{Failure Modes}
The most common failure modes are (1) in heavily forested areas (\cref{fig:supp_astroloc_Amazon}), which share similar visual characteristics to other areas in the same (as well as nearby) forests; (2) in coastal regions where very little land is in the image, causing many mostly water database images to be retrieved (\cref{fig:supp_vinsat}, bottom right); and (3) the presence of occlusion, typically in the form of clouds (\cref{fig:supp_vinsat}, many examples). Training specifically to ignore cloud regions may improve performance in the case of occlusions. Database curation, removing images with minimal landmass, can help alleviate the mostly water retrieval results. Further hard negative mining may help disambiguate similar forested regions, though we note that in practice these similar (but not the same) predictions are typically filtered by a local feature-based verification method like EarthMatch \cite{Berton_2024_EarthMatch}.Overcoming these failure modes will be explored in future work.

{
    \small
    \bibliographystyle{ieeenat_fullname}
    \bibliography{main}

\begin{thebibliography}{43}
\providecommand{\natexlab}[1]{#1}
\providecommand{\url}[1]{\texttt{#1}}
\expandafter\ifx\csname urlstyle\endcsname\relax
  \providecommand{\doi}[1]{doi: #1}\else
  \providecommand{\doi}{doi: \begingroup \urlstyle{rm}\Url}\fi

\bibitem[Ali-Bey et~al.(2022)Ali-Bey, Chaib-draa, and Giguere]{AliBey_2022_BMVC}
Amar Ali-Bey, Brahim Chaib-draa, and Philippe Giguere.
\newblock Global proxy-based hard mining for visual place recognition.
\newblock In \emph{33rd British Machine Vision Conference 2022, {BMVC} 2022, London, UK, November 21-24, 2022}. {BMVA} Press, 2022.

\bibitem[Ali-bey et~al.(2022)Ali-bey, Chaib-draa, and Gigu{\`e}re]{Alibey_2022_gsvcities}
Amar Ali-bey, Brahim Chaib-draa, and Philippe Gigu{\`e}re.
\newblock Gsv-cities: Toward appropriate supervised visual place recognition.
\newblock \emph{Neurocomputing}, 513:\penalty0 194--203, 2022.

\bibitem[{Arandjelović} et~al.(2018){Arandjelović}, Gronat, Torii, Pajdla, and Sivic]{Arandjelovic_2018_netvlad}
Relja {Arandjelović}, Petr Gronat, Akihiko Torii, Tomas Pajdla, and Josef Sivic.
\newblock {NetVLAD}: {CNN} architecture for weakly supervised place recognition.
\newblock \emph{IEEE Transactions on Pattern Analysis and Machine Intelligence}, 40\penalty0 (6):\penalty0 1437--1451, 2018.

\bibitem[Ayush et~al.(2021)Ayush, Uzkent, Meng, Tanmay, Burke, Lobell, and Ermon]{Ayush_2021_geography}
Kumar Ayush, Burak Uzkent, Chenlin Meng, Kumar Tanmay, Marshall Burke, David Lobell, and Stefano Ermon.
\newblock Geography-aware self-supervised learning.
\newblock \emph{IEEE International Conference on Computer Vision}, 2021.

\bibitem[Berton et~al.(2024{\natexlab{a}})Berton, Goletto, Trivigno, Stoken, Caputo, and Masone]{Berton_2024_EarthMatch}
Gabriele Berton, Gabriele Goletto, Gabriele Trivigno, Alex Stoken, Barbara Caputo, and Carlo Masone.
\newblock Earthmatch: Iterative coregistration for fine-grained localization of astronaut photography.
\newblock In \emph{Proceedings of the IEEE/CVF Conference on Computer Vision and Pattern Recognition (CVPR) Workshops}, 2024{\natexlab{a}}.

\bibitem[Berton et~al.(2024{\natexlab{b}})Berton, Stoken, Caputo, and Masone]{Berton_2024_EarthLoc}
Gabriele Berton, Alex Stoken, Barbara Caputo, and Carlo Masone.
\newblock Earthloc: Astronaut photography localization by indexing earth from space.
\newblock In \emph{IEEE Conference on Computer Vision and Pattern Recognition}, 2024{\natexlab{b}}.

\bibitem[Bianchi and Barfoot(2021)]{bianchi2021uav}
Mollie Bianchi and Timothy~D. Barfoot.
\newblock Uav localization using autoencoded satellite images.
\newblock \emph{arXiv preprint arXiv:2102.05692}, 2021.

\bibitem[Dai et~al.(2022)Dai, Zheng, Feng, Chen, and Yang]{dai2022finding}
Ming Dai, Enhui Zheng, Zhenhua Feng, Jiahao Chen, and Wankou Yang.
\newblock Finding point with image: A simple and efficient method for uav self-localization.
\newblock \emph{arXiv preprint arXiv:2208.06561}, 2022.

\bibitem[de~Miguel et~al.(2022)de~Miguel, Bennie, Rosenfeld, Dzurjak, and Gaston]{Sanchez_2022_artificial_lightning}
Alejandro~Sánchez de Miguel, Jonathan Bennie, Emma Rosenfeld, Simon Dzurjak, and Kevin~J. Gaston.
\newblock Environmental risks from artificial nighttime lighting widespread and increasing across europe.
\newblock \emph{Science Advances}, 8\penalty0 (37):\penalty0 eabl6891, 2022.

\bibitem[DeTone et~al.(2018)DeTone, Malisiewicz, and Rabinovich]{Detone_2018_superpoint}
Daniel DeTone, Tomasz Malisiewicz, and Andrew Rabinovich.
\newblock Superpoint: Self-supervised interest point detection and description.
\newblock In \emph{Proceedings of the IEEE conference on computer vision and pattern recognition workshops}, pages 224--236, 2018.

\bibitem[Deuser et~al.(2023)Deuser, Habel, Werner, and Oswald]{Deuser_2023_ogcl_uav_view}
Fabian Deuser, Konrad Habel, Martin Werner, and Norbert Oswald.
\newblock Orientation-guided contrastive learning for uav-view geo-localisation.
\newblock In \emph{Proceedings of the 2023 Workshop on UAVs in Multimedia: Capturing the World from a New Perspective}, page 7–11, New York, NY, USA, 2023. Association for Computing Machinery.

\bibitem[Dosovitskiy et~al.(2021)Dosovitskiy, Beyer, Kolesnikov, Weissenborn, Zhai, Unterthiner, Dehghani, Minderer, Heigold, Gelly, Uszkoreit, and Houlsby]{Dosovitskiy_2021_vit}
Alexey Dosovitskiy, Lucas Beyer, Alexander Kolesnikov, Dirk Weissenborn, Xiaohua Zhai, Thomas Unterthiner, Mostafa Dehghani, Matthias Minderer, Georg Heigold, Sylvain Gelly, Jakob Uszkoreit, and Neil Houlsby.
\newblock {An Image is Worth 16x16 Words: Transformers for Image Recognition at Scale}.
\newblock \emph{ArXiv}, abs/2010.11929, 2021.

\bibitem[Douze et~al.(2024)Douze, Guzhva, Deng, Johnson, Szilvasy, Mazaré, Lomeli, Hosseini, and Jégou]{Douze_2024_faiss}
Matthijs Douze, Alexandr Guzhva, Chengqi Deng, Jeff Johnson, Gergely Szilvasy, Pierre-Emmanuel Mazaré, Maria Lomeli, Lucas Hosseini, and Hervé Jégou.
\newblock The faiss library.
\newblock \emph{arXiv}, 2024.

\bibitem[Gaston and S\'{a}nchez~de Miguel(2022)]{Gaston_2022_environ_impacts_artificial_light}
Kevin~J. Gaston and Alejandro S\'{a}nchez~de Miguel.
\newblock Environmental impacts of artificial light at night.
\newblock \emph{Annual Review of Environment and Resources}, 47\penalty0 (1):\penalty0 373--398, 2022.

\bibitem[Ge et~al.(2020)Ge, Wang, Zhu, Zhao, and Li]{Ge_2020_sfrs}
Yixiao Ge, Haibo Wang, Feng Zhu, Rui Zhao, and Hongsheng Li.
\newblock Self-supervising fine-grained region similarities for large-scale image localization.
\newblock In \emph{Computer Vision -- ECCV 2020}, pages 369--386, Cham, 2020. Springer International Publishing.

\bibitem[He et~al.(2023)He, Cisneros, Keetha, Patrikar, Ye, Higgins, Hu, Kapoor, and Scherer]{He_2023_foundloc}
Yao He, Ivan Cisneros, Nikhil Keetha, Jay Patrikar, Zelin Ye, Ian Higgins, Yaoyu Hu, Parv Kapoor, and Sebastian Scherer.
\newblock Foundloc: Vision-based onboard aerial localization in the wild.
\newblock \emph{arXiv preprint arXiv:2310.16299}, 2023.

\bibitem[Izquierdo and Civera(2024)]{Izquierdo_2024_SALAD}
Sergio Izquierdo and Javier Civera.
\newblock Optimal transport aggregation for visual place recognition.
\newblock In \emph{IEEE Conference on Computer Vision and Pattern Recognition}, 2024.

\bibitem[Jehl et~al.(2013)Jehl, Farges, and Blanc]{sprites}
Augustin Jehl, Thomas Farges, and Elisabeth Blanc.
\newblock Color pictures of sprites from non-dedicated observation on board the international space station.
\newblock \emph{Journal of Geophysical Research: Space Physics}, 118\penalty0 (1):\penalty0 454--461, 2013.

\bibitem[Jégou et~al.(2011)Jégou, Douze, and Schmid]{Jegou_2011_productQ}
Hervé Jégou, Matthijs Douze, and Cordelia Schmid.
\newblock Product quantization for nearest neighbor search.
\newblock \emph{IEEE Trans. Pattern Anal. Mach. Intell.}, 33\penalty0 (1):\penalty0 117--128, 2011.

\bibitem[Keetha et~al.(2023)Keetha, Mishra, Karhade, Jatavallabhula, Scherer, Krishna, and Garg]{Keetha_2023_AnyLoc}
Nikhil Keetha, Avneesh Mishra, Jay Karhade, Krishna~Murthy Jatavallabhula, Sebastian Scherer, Madhava Krishna, and Sourav Garg.
\newblock Anyloc: Towards universal visual place recognition.
\newblock \emph{arXiv}, 2023.

\bibitem[Li et~al.(2024)Li, Liu, Qiu, and Li]{li2024transformer}
Shishen Li, Cuiwei Liu, Huaijun Qiu, and Zhaokui Li.
\newblock A transformer-based adaptive semantic aggregation method for uav visual geo-localization.
\newblock \emph{arXiv preprint arXiv:2401.01574}, 2024.

\bibitem[Lindenberger et~al.(2023)Lindenberger, Sarlin, and Pollefeys]{Lindenberger_2023_lightglue}
Philipp Lindenberger, Paul-Edouard Sarlin, and Marc Pollefeys.
\newblock {LightGlue: Local Feature Matching at Light Speed}.
\newblock In \emph{ICCV}, 2023.

\bibitem[McCleary et~al.(2024)McCleary, Gurumurthy, Fisch, Tayal, Manchester, and Lucia]{McCleary_2024_vinsat}
Kyle McCleary, Swaminathan Gurumurthy, Paulo~RM Fisch, Saral Tayal, Zachary Manchester, and Brandon Lucia.
\newblock Vinsat: Solving the lost-in-space problem with visual-inertial navigation.
\newblock In \emph{IEEE International Conference on Robotics and Automation (ICRA)}, 2024.

\bibitem[Musgrave et~al.(2020{\natexlab{a}})Musgrave, Belongie, and Lim]{Musgrave_2020_ML_reality}
Kevin Musgrave, Serge Belongie, and Ser-Nam Lim.
\newblock A metric learning reality check.
\newblock In \emph{Computer Vision -- ECCV 2020}, pages 681--699, Cham, 2020{\natexlab{a}}. Springer International Publishing.

\bibitem[Musgrave et~al.(2020{\natexlab{b}})Musgrave, Belongie, and Lim]{Musgrave_2020_PyTorchML}
Kevin Musgrave, Serge~J. Belongie, and Ser-Nam Lim.
\newblock Pytorch metric learning.
\newblock \emph{ArXiv}, abs/2008.09164, 2020{\natexlab{b}}.

\bibitem[{Nathani} et~al.(2022){Nathani}, {Iyer}, {Wang}, {Chitari}, {Wilson}, and {Norris}]{astro_photos_climate_patterns}
Aadil {Nathani}, Rishi {Iyer}, Annabelle {Wang}, Aarnav {Chitari}, Adele {Wilson}, and Hannah {Norris}.
\newblock {Observing Earth From Space: Using Astronaut Photography to Analyze Geographical Climate Patterns}.
\newblock In \emph{AGU Fall Meeting Abstracts}, pages ED44C--06, 2022.

\bibitem[{OpenStreetMap Contributors}(2023)]{osm_zoom_levels}
{OpenStreetMap Contributors}.
\newblock Zoom levels.
\newblock \url{https://wiki.openstreetmap.org/wiki/Zoom_levels}, 2023.

\bibitem[Oquab et~al.(2023)Oquab, Darcet, Moutakanni, Vo, Szafraniec, Khalidov, Fernandez, Haziza, Massa, El-Nouby, Howes, Huang, Xu, Sharma, Li, Galuba, Rabbat, Assran, Ballas, Synnaeve, Misra, Jegou, Mairal, Labatut, Joulin, and Bojanowski]{Oquab_2023_dinov2}
Maxime Oquab, Timothée Darcet, Theo Moutakanni, Huy~V. Vo, Marc Szafraniec, Vasil Khalidov, Pierre Fernandez, Daniel Haziza, Francisco Massa, Alaaeldin El-Nouby, Russell Howes, Po-Yao Huang, Hu Xu, Vasu Sharma, Shang-Wen Li, Wojciech Galuba, Mike Rabbat, Mido Assran, Nicolas Ballas, Gabriel Synnaeve, Ishan Misra, Herve Jegou, Julien Mairal, Patrick Labatut, Armand Joulin, and Piotr Bojanowski.
\newblock Dinov2: Learning robust visual features without supervision, 2023.

\bibitem[Schirmer et~al.(2019)Schirmer, Gallemore, Liu, Magle, DiNello, Ahmed, and Gilday]{Schirmer_2019_nightlight_behavior}
Aaron Schirmer, Caleb Gallemore, Ting Liu, Seth Magle, Elisabeth DiNello, Humerah Ahmed, and Thomas Gilday.
\newblock Mapping behaviorally relevant light pollution levels to improve urban habitat planning.
\newblock \emph{Scientific Reports}, 9, 2019.

\bibitem[Shockley and Bettinger(2021)]{Shockley_2021_sat_loc}
Liberty~M. Shockley and Robert~A. Bettinger.
\newblock Real-time aerospace vehicle position estimation using terrestrial illumination matching.
\newblock In \emph{IEEE 8th International Workshop on Metrology for AeroSpace (MetroAeroSpace),}, 2021.

\bibitem[Skinner et~al.(2022)Skinner, Coletti, Voss, Svitek, Lee, Auman, Patel, and Moyer]{Skinner_2022_cubesat_radar}
Mark~A. Skinner, Michael Coletti, Matthew~C. Voss, Tomas Svitek, John~C. Lee, Kerstyn Auman, Hemanshu Patel, and Eamonn~J. Moyer.
\newblock Mitigating cubesat confusion: Results of in-flight technical demonstrations of candidate tracking and identification technologies.
\newblock \emph{Journal of Space Safety Engineering}, 9\penalty0 (3):\penalty0 403--409, 2022.

\bibitem[Small(2022)]{Small_2022_spectrometry_urban_lightscape}
Christopher Small.
\newblock Spectrometry of the urban lightscape.
\newblock \emph{Technologies}, 10\penalty0 (4), 2022.

\bibitem[Stefanov and Evans(2015)]{IDC_Stefanov}
W.~L. Stefanov and C.~A. Evans.
\newblock Data collection for disaster response from the international space station.
\newblock \emph{The International Archives of the Photogrammetry, Remote Sensing and Spatial Information Sciences}, XL-7/W3:\penalty0 851--855, 2015.

\bibitem[Stewart et~al.(2022)Stewart, Robinson, Corley, Ortiz, Lavista~Ferres, and Banerjee]{Stewart_2022_TorchGeo}
Adam~J. Stewart, Caleb Robinson, Isaac~A. Corley, Anthony Ortiz, Juan~M. Lavista~Ferres, and Arindam Banerjee.
\newblock {TorchGeo}: Deep learning with geospatial data.
\newblock In \emph{Proceedings of the 30th International Conference on Advances in Geographic Information Systems}, pages 1--12, Seattle, Washington, 2022. Association for Computing Machinery.

\bibitem[Stoken and Fisher(2023)]{Stoken_2023_CVPR}
Alex Stoken and Kenton Fisher.
\newblock Find my astronaut photo: Automated localization and georectification of astronaut photography.
\newblock In \emph{CVPRW}, pages 6196--6205, 2023.

\bibitem[Stoken et~al.(2024)Stoken, Ilhardt, Lambert, and Fisher]{Stoken_2024_CVPR}
Alex Stoken, Peter Ilhardt, Mark Lambert, and Kenton Fisher.
\newblock (street) lights will guide you: Georeferencing nighttime astronaut photography of earth.
\newblock In \emph{Proceedings of the IEEE/CVF Conference on Computer Vision and Pattern Recognition (CVPR) Workshops}, pages 492--501, 2024.

\bibitem[Straub and Christian(2015)]{Straub_2015_sat_loc}
Miranda Straub and John~A. Christian.
\newblock Autonomous optical navigation for earth-observing satellites using coastline matching.
\newblock In \emph{AIAA Guidance, Navigation, and Control Conference}, 2015.

\bibitem[Trivigno et~al.(2024)Trivigno, Masone, Caputo, and Sattler]{Trivigno_2024_CVPR}
Gabriele Trivigno, Carlo Masone, Barbara Caputo, and Torsten Sattler.
\newblock The unreasonable effectiveness of pre-trained features for camera pose refinement.
\newblock In \emph{Proceedings of the IEEE/CVF Conference on Computer Vision and Pattern Recognition (CVPR)}, pages 12786--12798, 2024.

\bibitem[Wang et~al.(2019)Wang, Han, Huang, Dong, and Scott]{Wang_2019_multi_similarity_loss}
Xun Wang, Xintong Han, Weilin Huang, Dengke Dong, and Matthew~R Scott.
\newblock Multi-similarity loss with general pair weighting for deep metric learning.
\newblock In \emph{Proceedings of the IEEE Conference on Computer Vision and Pattern Recognition}, pages 5022--5030, 2019.

\bibitem[Yair et~al.(2023)Yair, Korman, Price, and Stibbe]{TLEs}
Yoav Yair, Melody Korman, Colin Price, and Eytan Stibbe.
\newblock Observing lightning and transient luminous events from the international space station during ilan-es: An astronaut's perspective.
\newblock \emph{Acta Astronautica}, 211:\penalty0 592--599, 2023.

\bibitem[Yao et~al.(2021)Yao, Zhang, Qiu, Pan, and Mei]{Yao_2021_seco}
Ting Yao, Yiheng Zhang, Zhaofan Qiu, Yingwei Pan, and Tao Mei.
\newblock Seco: Exploring sequence supervision for unsupervised representation learning.
\newblock In \emph{AAAI}, 2021.

\bibitem[Zheng et~al.(2023)Zheng, Shi, Wang, Liu, Fang, Wei, and Chua]{Zheng_2023_UAV_workshop}
Zhedong Zheng, Yujiao Shi, Tingyu Wang, Jun Liu, Jianwu Fang, Yunchao Wei, and Tat-seng Chua.
\newblock Uavm '23: 2023 workshop on uavs in multimedia: Capturing the world from a new perspective.
\newblock In \emph{ACM MM}, page 9715–9717, New York, NY, USA, 2023. Association for Computing Machinery.

\bibitem[Zhu et~al.(2023)Zhu, Yang, Zhang, Wu, Yin, and Zhang]{Zhu_2023_uav_backbone_winnerUAV_mbeg}
Runzhe Zhu, Mingze Yang, Kaiyu Zhang, Fei Wu, Ling Yin, and Yujin Zhang.
\newblock Modern backbone for efficient geo-localization.
\newblock In \emph{Proceedings of the 2023 Workshop on UAVs in Multimedia: Capturing the World from a New Perspective}, page 31–37, New York, NY, USA, 2023. Association for Computing Machinery.

\end{thebibliography}
}

\end{document}